\documentclass{article}

\usepackage{microtype}
\usepackage{graphicx}
\usepackage{subcaption}
\usepackage{booktabs} 

\usepackage{hyperref}

\usepackage{paralist}       
\usepackage{multirow}       
\usepackage{xcolor}         
\usepackage{colortbl}       
\usepackage{array}          
\usepackage{arydshln}       
\usepackage{xspace}         
\usepackage{url}            
\usepackage{enumitem}       
\usepackage{tcolorbox}      
\usepackage{setspace}       
\tcbuselibrary{breakable}   

\usepackage[preprint]{icml2026}

\usepackage{amsmath}
\usepackage{amssymb}
\usepackage{mathtools}
\usepackage{amsthm}
\usepackage{bm}             

\usepackage[capitalize,noabbrev]{cleveref}

\theoremstyle{plain}

\theoremstyle{definition}

\theoremstyle{remark}

\usepackage[textsize=tiny]{todonotes}

\newcolumntype{B}{>{\columncolor{blue!4}}c}
\newcolumntype{d}{>{\columncolor{brown!4}}c}
\newcolumntype{R}{>{\columncolor{red!4}}c}

\makeatletter
\def\adl@drawiv#1#2#3{%
        \hskip.5\tabcolsep
        \xleaders#3{#2.5\@tempdimb #1{1}#2.5\@tempdimb}%
                #2\z@ plus1fil minus1fil\relax
        \hskip.5\tabcolsep}
\newcommand{\cdashlinelr}[1]{%
  \noalign{\vskip\aboverulesep
            \global\let\@dashdrawstore\adl@draw
            \global\let\adl@draw\adl@drawiv}
  \cdashline{#1}
  \noalign{\global\let\adl@draw\@dashdrawstore
            \vskip\belowrulesep}}
\makeatother

\newcommand{\ours}{\textsc{Pns}}

\icmltitlerunning{Not All Negative Samples Are Equal: LLMs Learn Better from Plausible Reasoning}

\begin{document}

\twocolumn[
  \icmltitle{Not All Negative Samples Are Equal: LLMs Learn Better from Plausible Reasoning}



  \icmlsetsymbol{equal}{*}

  \begin{icmlauthorlist}
    \icmlauthor{Zixiang Di}{equal,ecnu}
    \icmlauthor{Jinyi Han}{equal,ecnu}
    \icmlauthor{Shuo Zhang}{ecnu}
    \icmlauthor{Ying Liao}{fdu}\\
    \icmlauthor{Zhi Li}{inde}
    \icmlauthor{Xiaofeng Ji}{inde}
    \icmlauthor{Yongqi Wang}{inde}
    \icmlauthor{Zheming Yang}{inde}
    \icmlauthor{Ming Gao}{ecnu}\\
    \icmlauthor{Bingdong Li\textsuperscript{\textdagger}}{ecnu}
    \icmlauthor{Jie Wang\textsuperscript{\textdagger}}{inde}
  \end{icmlauthorlist}

  \icmlaffiliation{ecnu}{East China Normal University}
  \icmlaffiliation{fdu}{Fudan University}
  \icmlaffiliation{inde}{Independent Researcher}

  \icmlequalcontribution{Zixiang Di}{51265901113@stu.ecnu.edu.cn}
  \icmlequalcontribution{Jinyi Han}{jinyihan099@gmail.com}

  \icmlcorrespondingauthor{Bingdong Li}{bdli@cs.ecnu.edu.cn}
  \icmlcorrespondingauthor{Jie Wang}{cyrusmayhaohao@gmail.com}

  \icmlkeywords{Machine Learning, ICML}

  \vskip 0.3in
]



\printAffiliationsAndNotice{}  

\begin{abstract}
Learning from negative samples holds great promise for improving Large Language Model (LLM) reasoning capability, yet existing methods treat all incorrect responses as equally informative, overlooking the crucial role of sample quality. To address this, we propose \textbf{Plausible Negative Samples} (\ours{}), a method that synthesizes high-quality negative samples exhibiting expected format and structural coherence while ultimately yielding incorrect answers. \ours{} trains a dedicated model via reverse reinforcement learning (RL) guided by a composite reward combining format compliance, accuracy inversion, reward model assessment, and chain-of-thought evaluation, generating responses nearly indistinguishable from correct solutions. We further validate \ours{} as a plug-and-play data source for preference optimization across three backbone models on seven mathematical reasoning benchmarks. Results demonstrate that \ours{} consistently outperforms other negative sample synthesis methods, achieving an average improvement of 2.03\% over RL-trained models.
\end{abstract}

\section{Introduction}
Existing training paradigms for LLMs mainly rely on high-quality positive samples to improve ~\cite{ouyang2022training,LLM_wei2022finetuned}. While this approach has demonstrated remarkable success, it largely neglects the rich information contained in negative samples. Growing evidence shows that incorporating negative samples is not merely noise; instead, it can serve as critical counter-examples that reveal flawed reasoning patterns~\cite{zhu2025continual,hamdan2025how,tian2026learning}. Leveraging these signals help better shape decision boundaries, further enhancing reasoning capability~\cite{wang2024learning}.
\begin{figure}[t]
\centering
\includegraphics[width=\linewidth]{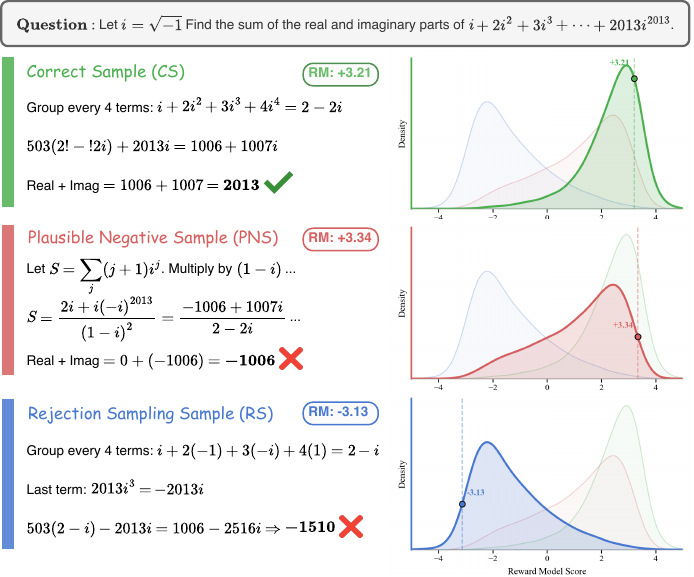}
\vspace{-0.4cm}
\caption{\textbf{Comparison of a correct sample (CS), a plausible negative sample (\ours{}), and a rejection sampling sample (RS).} \ours{} produces an incorrect answer through deceptively coherent reasoning, receiving a higher RM score ($+3.34$) than even the correct solution ($+3.21$), while RS is easily detected ($-3.13$).}
\label{fig:intro}
\vspace{-0.4cm}
\end{figure}

Despite the great potential of negative samples in LLM training, existing methods largely overlook their quality and treat all errors as equally informative. Techniques such as rejection sampling typically label a sample as negative solely based on the correctness of the final answer~\cite{khaki-etal-2024-rs,LLM_xiong2025minimalist,LLM_quamar2025stars}. They ignore the fact that different errors carry fundamentally distinct learning values~\cite{LLM_luo2024omegaprm,zhang-etal-2025-lessons}. For example, a trivial formatting mistake is treated no differently than a deeply flawed reasoning chain. While models quickly learn to avoid obvious or superficial errors, they fail to benifit from the high-value superivison hidden in subtle reasoning mistakes.

Inspired by the error-driven learning in humans~\cite{LLM_COG_holroyd2002ern}, simple mistakes are easy to recognize and correct, whereas meaningful cognitive progress comes from addressing errors that appear plausible. The same principle applies to LLMs. Obvious logical violations mainly help the model establish coarse decision boundaries, offering limited room for further optimization~\cite{LLM_kumar2010selfpaced}. In contrast, errors that preserve a coherent reasoning structure yet deviate at critical intermediate steps offer richer and more informative learning signals. These ``plausible but wrong'' samples lie close to correct solutions, making them difficult to distinguish and highly effective at exposing subtle weaknesses in reasoning, as shown in Figure \ref{fig:intro}.

However, generating high-quality negative samples is fundamentally difficult because it involves an inherent conflict between correctness and plausibility. On the one hand, the reasoning process must appear coherent, structured, and convincing; on the other hand, the final conclusion must be wrong. Yet in LLMs, strong reasoning naturally drives the model toward correct answers, making it highly unstable to preserve plausible reasoning while enforcing an incorrect outcome. Moreover, errors cannot be injected by simply breaking the logical structure, since such artifacts are easily detected by reward models. Instead, the mistake must be locally plausible but globally wrong, meaning that each individual step appears reasonable while their composition leads to an incorrect conclusion.

To address this, we propose the \ours{} generation method, designed to synthesize high-quality negative samples. These samples feature reasoning paths that mimic the expected format and exhibit structural coherence, yet ultimately yield incorrect answers. \ours{} is built upon a Reverse Group Relative Policy Optimization (Reverse GRPO) algorithm~\cite{LLM_shao2024deepseekmath} and leverages a hybrid reward mechanism to steer the model toward generating these structured yet erroneous trajectories. We further validate the efficacy of these samples within the Direct Preference Optimization (DPO)~\cite{rafailov2023direct} framework. Experimental results demonstrate that even models previously refined via Reinforcement Learning with Verifiable Rewards (RLVR)~\cite{wen2025rlvr_reasoning,liu2025rise} significantly from \ours{}-derived data; notably, Qwen2.5-7B-Instruct achieves an average improvement of 2.03\% across seveal mathematical benchmarks.

To summarize, our contributions are three-fold:
\begin{itemize}[wide=1pt, itemsep=1pt]
    \item To generate high-quality negative samples that are nearly indistinguishable from correct solutions, we propose \ours{}, which leverages reverse reinforcement learning to synthesize negatives with coherent reasoning processes and expected format while arriving at incorrect answers.
    \item We design the \ours{} Reward, a composite signal combining format compliance, accuracy inversion, reward model assessment, and chain-of-thought evaluation to improve reasoning quality while enforcing answer incorrectness.
    \item Experiments across three backbone models and seven benchmarks demonstrate that negative samples synthesized by \ours{} can further improve performance on top of RL-refined models.
\end{itemize}



\section{Related Work}
\subsection{Reinforcement Learning with Verifiable Rewards}
\label{subsec:rlvr}

RL has demonstrated significant potential in enhancing the reasoning capabilities of LLMs~\citep{ouyang2022training}.
Recently, Reinforcement Learning with Verifiable Rewards (RLVR) has emerged as a dominant paradigm, replacing human feedback with automatically checkable objectives such as mathematical verification~\citep{shao2024deepseekmath}, leading to the development of large reasoning models (LRMs) including OpenAI-o1~\citep{jaech2024openai}, DeepSeek-R1~\citep{guo2025deepseek}, and Kimi-k1.5~\citep{team2025kimi}, which leverage long chain-of-thought (CoT) reasoning through test-time scaling~\citep{zeng2025simplerl, hu2025openreasonerzero, yuan2025kardia}.
Building on the foundational GRPO algorithm~\citep{shao2024deepseekmath}, several improvements have been proposed: DAPO~\citep{yu2025dapo} introduces decoupled clipping and dynamic sampling; process reward models~\citep{yuan2025free, cui2025prime} provide denser supervision signals; and recent analysis~\citep{yue2025does} reveals that RLVR primarily redistributes probability mass over existing reasoning paths rather than expanding the model's fundamental reasoning capacity.
These advances focus exclusively on guiding models toward correct answers. In contrast, our work reverses the reward signal to generate plausible-but-incorrect reasoning, and then leverages these samples via preference optimization to further strengthen model performance.

\vspace{-0.2cm}

\subsection{Learning from Negative Samples}
\label{subsec:negative}

Prior work leverages negative samples to improve LLMs mainly in three ways: prompting, fine-tuning, and reinforcement learning.
Prompt-based methods place negative examples in the input context to steer generation away from undesired behaviors~\citep{gao2024contrastive, alazraki-etal-2025-need}. However, their effectiveness depends on the model's existing instruction-following and reasoning abilities, limiting their impact on weaker models.
Fine-tuning-based approaches incorporate negative data more directly. A common strategy converts incorrect trajectories into positive CoT supervision via teacher rewriting or error correction~\citep{an2023learning, pan2025lemma, yu2025selferror}. Other work explicitly marks negative samples during training via mistake tuning~\citep{tong2024can} or trajectory-level prefixes~\citep{wang2024learning}.
Recently, \citet{tian2026learning} show that directly training on negative reasoning trajectories can improve out-of-domain generalization, and \citet{hamdan2025how} find that near-miss negatives provide substantially stronger learning signals than random incorrect ones.
These results underscore that the quality of negative samples is critical for effective learning. To this end, we propose reverse GRPO, which trains a dedicated model to generate negative samples with plausible reasoning processes, and uses these samples to further enhance model performance.

\begin{figure*}[t]
\centering
\includegraphics[width=0.7\linewidth]{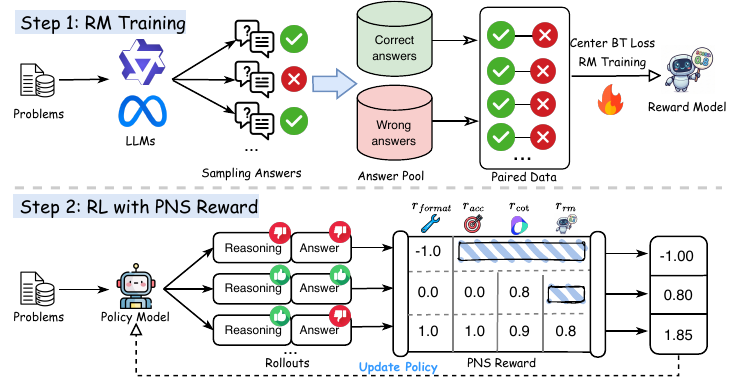}
\vspace{-0.1cm}
\caption{\textbf{Overview of plausible negative sample synthesis.} \textbf{Step 1:} We construct preference pairs from sampled responses and train a Reward Model to assess reasoning quality. \textbf{Step 2:} We optimize a Policy Model via RL with the \ours{} Reward to generate responses with high reasoning quality but incorrect answers.}
\label{fig:overview}
\vspace{-0.2cm}
\end{figure*}

\section{Method}

We propose Plausible Negative Samples (\ours{}), a method that synthesizes high-quality negative samples with coherent reasoning but incorrect answers. As shown in Figure~\ref{fig:overview}, \ours{} consists of two phases. In \textbf{Phase 1} (\S\ref{subsec:synthesizing}), we train a Reward Model on preference pairs constructed from correct and incorrect responses (Step 1), then incorporate it into the \ours{} Reward to optimize a Policy Model via RL (Step 2), steering it to generate responses with high reasoning quality but incorrect answers. In \textbf{Phase 2} (\S\ref{subsec:learning}), the trained \ours{} Model serves as a plug-and-play data generator, providing plausible negative samples as rejected data for preference optimization on the target LLM.

\subsection{Synthesizing Plausible Negative Samples}
\label{subsec:synthesizing}

To generate negative samples that are not merely wrong but convincingly wrong, we train a \ours{} Model based on Qwen2.5-7B-Instruct~\citep{qwen2025qwen25technicalreport} through a two-step process. We first build a reward model capable of assessing reasoning quality, then employ it as a key component in our reverse reinforcement learning framework.

\subsubsection{Step 1: RM for Response Quality Assessment}
\label{subsubsec:rm}

We train a scalar reward model (RM) based on Qwen3-4B~\citep{yang2025qwen3} to provide fine-grained quality assessments of mathematical reasoning responses. The RM assigns high scores to responses with correct derivations and answers, and low scores to those containing errors.

We adopt the Bradley-Terry loss~\citep{bradley1952rank} as the training objective. Inspired by reward model training practices in~\citet{touvron2023llama2}, we augment the standard loss with an $L_2$ regularization term that encourages the model's reward outputs to be symmetrically distributed around zero. We term the resulting objective the Center-Regularized Bradley-Terry (Center BT) Loss:
\begin{equation}
\resizebox{.9\linewidth}{!}{$\displaystyle
\mathcal{L}_{\text{Center BT}} = -\log\Big(\sigma\big(r(y_w) - r(y_l)\big)\Big) + \lambda \big(r(y_w) + r(y_l)\big)^2,
\label{eq:bt_loss}
$}
\end{equation}
where $y_w$ and $y_l$ denote the preferred and dispreferred responses respectively, $r(\cdot)$ is the scalar reward predicted by the RM, $\sigma(x) = 1 / (1 + e^{-x})$ is the sigmoid function, and $\lambda$ is the regularization coefficient. The first term maximizes the probability of the RM assigning a higher score to the preferred response, while the second term penalizes deviations of the reward sum from zero, encouraging a symmetric score distribution centered at the origin. This regularization improves the separability between high- and low-quality responses, as validated by the well-separated reward distributions shown in Figure~\ref{fig:rm_distribution}.

\subsubsection{Step 2: RL with \ours{} Reward}
\label{subsubsec:rl}

This step constitutes the core of our method. We design the \textit{\ours{} Reward} that inverts the conventional RL reward paradigm: instead of rewarding correct answers, we encourage the model to produce responses whose final answers are \textit{incorrect} yet whose reasoning processes appear \textit{plausible}. We employ GRPO~\citep{shao2024deepseekmath} as the base RL algorithm. For each question $q$, the model samples a group of outputs $O = \{o_1, o_2, \ldots, o_G\}$, and the policy is optimized using group-relative advantages computed from the \ours{} Reward, which is composed of four complementary signals.

\begin{figure}[t]
\centering
\includegraphics[width=0.6\linewidth]{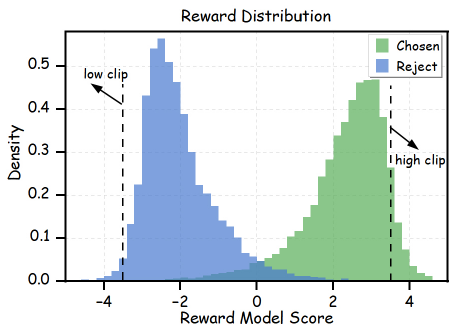}
\vspace{-0.1cm}
\caption{\textbf{Reward score distribution of our trained RM.} The RM achieves 98.89\% pairwise accuracy over $n{=}14{,}181$ pairs, with well-separated distributions for chosen (blue) and rejected (yellow) responses.}
\label{fig:rm_distribution}
\vspace{-0.2cm}
\end{figure}

\textbf{Hybrid Format Score.}
The format score $r_{\text{format}}(y)$ ensures that the model's output conforms to expected structural conventions, preventing reward hacking where the \ours{} Model deliberately violates format requirements to make answer extraction unreliable, thereby trivially producing incorrect outputs without genuine reasoning.
We adopt a hybrid approach combining rule-based and LLM-as-a-Judge evaluation.

\textit{Rule-based Constraints.} Let $C_i(y) \in \{0, 1\}$ indicate whether output $y$ satisfies constraint $C_i$. We define five structural constraints:\\
$\bm{C_1}$: The response contains exactly one \texttt{<think>} and one \texttt{</think>} tag;\\
$\bm{C_2}$: The content before \texttt{</think>} is non-empty;\\
$\bm{C_3}$: The content after \texttt{</think>} is non-empty;\\
$\bm{C_4}$: The last \texttt{\textbackslash boxed\{\ldots\}} expression appears after \texttt{</think>};\\
$\bm{C_5}$: The content within the final boxed expression is non-empty.\\
The rule-based score is defined as:
\begin{equation}
r_{\text{rule}}(y) = \prod_{i=1}^{5} C_i(y),
\label{eq:rule_score}
\end{equation}
which assigns a nonzero score only to outputs satisfying all structural requirements.

\textit{LLM-as-a-Judge.} Beyond structural validity, we employ an external judge model to verify that the response contains clear, verifiable mathematical derivations or computations. The judge outputs a binary score $r_{\text{judge}}(y) \in \{0, 1\}$ (see Appendix~\ref{sec:judge_prompt} for the detailed prompt).

The final format score combines both components:
\begin{equation}
r_{\text{format}}(y) = r_{\text{rule}}(y) \cdot r_{\text{judge}}(y).
\label{eq:format_score}
\end{equation}
Only responses satisfying both rule-based constraints and the LLM judge receive a nonzero format score.

\textbf{Accuracy Score.}
The accuracy score $r_{\text{acc}}(y)$ is a binary indicator that determines whether the response's final answer matches the ground truth:
\begin{equation}
r_{\text{acc}}(y) = \begin{cases}
1, & \text{if } y = a^*, \\
0, & \text{otherwise},
\end{cases}
\label{eq:acc_score}
\end{equation}
where $a^*$ denotes the ground-truth answer. This score is central to our reverse reward design: responses with $r_{\text{acc}} = 0$ (incorrect answers) are candidates for plausible negatives, while those with $r_{\text{acc}} = 1$ (correct answers) receive neutral rewards to prevent their reinforcement.

\textbf{RM Score.}
The reward model score $r_{\text{rm}}(y)$ serves as an expert assessment of reasoning quality, guiding the \ours{} Model toward generating responses that are convincing to quality evaluators despite being factually incorrect.
To enhance discriminability across responses and mitigate RM hacking in late-stage training, we apply clipping and bucketing to the raw RM output, followed by normalization to $[0, 1]$:
\begin{equation}
r_{\text{rm}}(y) = \frac{\mathrm{Bucket}\!\big(\mathrm{Clip}(r(y),\, s_{\min},\, s_{\max})\big) - s_{\min}}{s_{\max} - s_{\min}},
\label{eq:rm_score}
\end{equation}
where $\mathrm{Clip}(\cdot)$ clamps the raw score to $[s_{\min}, s_{\max}]$ and $\mathrm{Bucket}(\cdot)$ maps the clipped value to the nearest boundary in a predefined set $\mathcal{B}$.
We determine the clipping range and bucket boundaries empirically: we sample 8 responses per training query from Qwen2.5-7B-Instruct and score all query-response pairs with the RM.
Based on the resulting distribution (Figure~\ref{fig:rm_distribution}), we set $s_{\min} = -3.5$, $s_{\max} = 3.5$, and $\mathcal{B} = \{-3.5, -3, -2.5, -2, -1, 0, 1, 2, 2.5, 3, 3.5\}$.

\textbf{CoT Score.}
The chain-of-thought score $r_{\text{cot}}(y)$ safeguards the baseline quality of the generated reasoning traces. Since reverse RL training may degrade the model's instruction-following capabilities, the CoT score evaluates four dimensions via an LLM-as-a-Judge. Let $s_i(y) \in \{0, 1, 2, 3\}$ denote the score for dimension $i$:\\
$\bm{D_1}$: \textit{Reasoning validity}, where \texttt{<think>} contains steps directly relevant to the problem;\\
$\bm{D_2}$: \textit{Reasoning-conclusion consistency}, where the reasoning steps logically support the stated conclusion with verifiable mathematical derivations;\\
$\bm{D_3}$: \textit{Instruction following}, where the response follows all task constraints without off-task content;\\
$\bm{D_4}$: \textit{Generation quality}, where the response does not exhibit mechanical repetition or unnecessary language mixing.\\
The CoT score aggregates all and normalizes to $[0, 1]$:

\vspace{-0.2cm}

\begin{equation}
r_{\text{cot}}(y) = \frac{1}{12} \sum_{i=1}^{4} s_i(y).
\label{eq:cot_score}
\end{equation}

\vspace{-0.2cm}

\textbf{\ours{} Reward.}
We compose the four scores into the final \ours{} Reward $r_{\text{PNS}}(y)$, whose structure reflects our core objective: rewarding responses that are \textit{format-compliant}, \textit{factually incorrect}, yet exhibit \textit{high reasoning quality}:
\begin{equation}
\resizebox{.9\linewidth}{!}{$\displaystyle
r_{\text{PNS}}(y) = \begin{cases}
1 + \lambda_r \cdot r_{\text{rm}}(y) + \lambda_c \cdot r_{\text{cot}}(y), & \text{if } r_{\text{format}} {=} 1 \land r_{\text{acc}} {=} 0, \\[3pt]
0 + \lambda_c \cdot r_{\text{cot}}(y), & \text{if } r_{\text{format}} {=} 1 \land r_{\text{acc}} {=} 1, \\[3pt]
-1, & \text{if } r_{\text{format}} {=} 0,
\end{cases}
$}
\label{eq:pns_reward}
\end{equation}
where $\lambda_r$ and $\lambda_c$ are weighting coefficients for the RM score and CoT score, respectively.
The first case assigns the highest reward to format-compliant responses with incorrect answers, with bonuses proportional to reasoning quality, steering the model toward generating plausible reasoning that nonetheless leads to wrong conclusions.
The second case assigns a small positive reward to correct answers based on CoT quality, maintaining reasonable generation quality without strongly reinforcing correct outputs.
The third case penalizes format violations, discouraging degenerate outputs.
This reversed reward structure trains the \ours{} Model to produce samples that maximally challenge downstream preference learning.

\subsection{Learning from Plausible Negative Samples}
\label{subsec:learning}

The plausible negative samples synthesized by the \ours{} Model constitute a versatile, plug-and-play training resource for enhancing model performance across different training stages.

In contrast to recent work that directly applies supervised fine-tuning on negative reasoning samples to improve out-of-domain generalization~\citep{tian2026learning}, our method incorporates plausible negative samples within a preference optimization framework.
Specifically, for each training query $q$, we sample responses from the target model and retain those with correct final answers as \textit{chosen} responses $y_w$. Concurrently, we sample from the trained \ours{} Model to obtain plausible negative samples as \textit{rejected} responses $y_l$. These chosen-rejected pairs are then used to train the target model via DPO~\citep{rafailov2023direct}:
\begin{equation}
\resizebox{.9\linewidth}{!}{$\displaystyle
\mathcal{L}_{\text{DPO}} = -\mathbb{E}_{(q, y_w, y_l)} \!\left[\log \sigma \!\left(\beta \!\left(\log \frac{\pi_\theta(y_w | q)}{\pi_{\text{ref}}(y_w | q)} - \log \frac{\pi_\theta(y_l | q)}{\pi_{\text{ref}}(y_l | q)}\right)\!\right)\!\right]\!,
\label{eq:dpo}
$}
\end{equation}
where $\pi_\theta$ is the policy being optimized, $\pi_{\text{ref}}$ is the reference policy, and $\beta$ controls the deviation from the reference.

This design offers two key advantages.
First, by employing plausible negatives as rejected data rather than trivially incorrect samples, the preference optimization is exposed to harder contrastive pairs that share similar reasoning patterns with correct solutions.
This forces the model to develop finer-grained discrimination between sound and subtly flawed reasoning, yielding more robust improvements than training with easily distinguishable negatives.
Second, the plug-and-play nature of the approach allows the synthesized plausible negatives to be seamlessly applied at different stages of the training pipeline, either to a base model directly or to a model that has already undergone reinforcement learning, consistently yielding performance gains on both in-domain and out-of-domain tasks.

\section{Experiments}

\subsection{Experiment Setups}
\label{subsec: experimental setup}

\textbf{Model} \quad
We conduct experiments on three widely used backbone models: Qwen2.5-7B-Instruct, Qwen2.5-3B-Instruct~\citep{qwen2025qwen25technicalreport}, and Llama3.1-8B-Instruct~\citep{dubey2024llama}. These models span different parameter scales and model families, allowing us to evaluate the generality of our approach. For the \ours{} Model, we uniformly use Qwen2.5-7B-Instruct as the base model and apply the synthesized plausible negative samples across all three backbone models. The scalar reward model is trained from Qwen3-4B~\citep{yang2025qwen3}.

\textbf{Training Datasets} \quad
 For the RM training and RL stage, we use DAPO-Math\footnote{\url{https://huggingface.co/datasets/BytedTsinghua-SIA/DAPO-Math-17k}} after filtering out all Chinese-language problems. The reward model is trained on preference pairs constructed by sampling multiple responses per training query and pairing correct and incorrect answers.

\textbf{Baselines} \quad
We compare \ours{} against several baselines:
\begin{inparaenum}[\it (1)]
\item \textbf{Base}, the original instruction-tuned model without additional training;
\item \textbf{RL}, applying GRPO~\citep{shao2024deepseekmath} to the base model;
\item \textbf{RL + Negative}, following~\citet{tian2026learning}, performing SFT on the RL model with negative reasoning trajectories to improve out-of-domain generalization;
\item \textbf{RL + Positive \& Negative (RS)}, applying DPO on the RL model with chosen data from the model's own correct responses and rejected data from rejection sampling;
\item \textbf{RL + Positive \& Negative (LLM Judge)}, applying DPO on the RL model with rejected data generated by a model trained with LLM-judge-based quality evaluation; and
\item \textbf{RL + Positive \& Negative (\ours{})}, applying DPO on the RL model using plausible negative samples synthesized by our \ours{} Model as rejected data.
\end{inparaenum}
Detailed hyperparameter settings are provided in Appendix~\ref{sec:exp_details}.

\subsection{Evaluation}
\label{subsec: evaluation datasets}

\textbf{Evaluation Datasets} \quad
We evaluate models on both in-domain and out-of-domain benchmarks.
\textbf{(i) In-domain:} We assess mathematical reasoning on MATH500~\citep{hendrycks2measuring}, AIME'24, AIME'25, AMC~\citep{li2024numinamath}, and Olympiad~\citep{he2024olympiadbench}.
\textbf{(ii) Out-of-domain:} We evaluate generalization capability on ARC~\citep{clark2018think} and GPQA-Diamond (GPQA)~\citep{rein2024gpqa}.

\textbf{Inference} \quad
For AIME'24, AIME'25, and AMC, we report \textit{avg@10} at temperature 0.6 to reduce variance on these competition-level benchmarks. For all other benchmarks, we report \textit{pass@1} at temperature 0 for deterministic evaluation.

\subsection{Main Results}
\label{subsec: main results}

\setlength\tabcolsep{5.7pt}
\begin{table*}[t]
\centering
\def\arraystretch{.99}
\setlength{\tabcolsep}{0.42em}
\resizebox{1.0\linewidth}{!}{
\begin{tabular}{ll BBBBB dd R}
\toprule
\multirow{2}{*}{\textbf{Model}} & \multirow{2}{*}{\textbf{Setting}}
& \multicolumn{5}{c}{\textbf{In-Domain}} & \multicolumn{2}{c}{\textbf{Out-of-Domain}} & \cellcolor{white}{\multirow{2}{*}{\textbf{AVG}}} \\
\cmidrule(lr){3-7} \cmidrule(lr){8-9}
& & \cellcolor{white}{MATH500} & \cellcolor{white}{AIME24} & \cellcolor{white}{AIME25} & \cellcolor{white}{Olympiad} & \cellcolor{white}{AMC} & \cellcolor{white}{ARC} & \cellcolor{white}{GPQA} \\
\midrule
\multirow{6}{*}{\textbf{Qwen2.5-7B-Instruct}}
& Base                         & 75.00 & 8.67  & 4.67  & 37.48 & 40.60 & 83.02 & 32.83 & 35.28 \\
& RL                           & 77.40 & 11.00 & 8.33  & 42.67 & 50.48 & 85.67 & 39.90 & 39.43 \\
\cdashlinelr{2-10}
& RL + Negative                & 76.20 & 8.33  & 9.00  & 39.85 & 46.39 & 86.26 & 36.87 & 37.86 \\
& RL + P\&N (RS)               & 77.20 & 10.67 & 8.33  & 41.33 & 51.57 & 85.92 & 35.86 & 38.86 \\
& RL + P\&N (LLM Judge)        & 76.80 & 10.33 & 6.33  & 40.44 & 51.33 & 88.40 & 35.86 & 38.69 \\
& RL + P\&N (\ours{})          & 77.80 & 14.67 & 9.67  & 40.89 & 49.28 & 90.44 & 36.36 & \textbf{39.89} \\
\midrule
\multirow{6}{*}{\textbf{Qwen2.5-3B-Instruct}}
& Base                         & 65.80 & 4.00  & 3.67  & 29.19 & 27.95 & 69.11 & 32.32 & 29.01 \\
& RL                           & 69.20 & 6.00  & 2.67  & 32.44 & 33.73 & 80.72 & 32.32 & 32.14 \\
\cdashlinelr{2-10}
& RL + Negative                & 67.20 & 5.67  & 3.00  & 31.70 & 32.05 & 79.61 & 35.86 & 31.89 \\
& RL + P\&N (RS)               & 63.40 & 4.67  & 1.33  & 29.33 & 30.48 & 75.85 & 21.21 & 28.28 \\
& RL + P\&N (LLM Judge)        & 69.00 & 5.67  & 2.67  & 31.41 & 32.53 & 81.74 & 32.83 & 31.98 \\
& RL + P\&N (\ours{})          & 69.40 & 6.67  & 4.00  & 31.85 & 33.61 & 81.14 & 34.85 & \textbf{32.69} \\
\midrule
\multirow{6}{*}{\textbf{Llama3.1-8B-Instruct}}
& Base                         & 42.60 & 3.00  & 1.33  & 15.70 & 13.49 & 56.74 & 24.75 & 19.70 \\
& RL                           & 49.40 & 6.33  & 0.33  & 16.89 & 22.65 & 69.80 & 30.81 & 24.53 \\
\cdashlinelr{2-10}
& RL + Negative                & 47.60 & 5.00  & 0.00  & 14.67 & 14.82 & 74.49 & 30.81 & 23.42 \\
& RL + P\&N (RS)               & 49.80 & 5.67  & 0.33  & 16.59 & 20.00 & 78.50 & 21.21 & 24.01 \\
& RL + P\&N (LLM Judge)        & 50.20 & 5.00  & 0.00  & 17.63 & 19.76 & 59.30 & 27.78 & 22.46 \\
& RL + P\&N (\ours{})          & 51.40 & 7.00  & 2.00  & 17.19 & 20.24 & 69.03 & 31.31 & \textbf{24.77} \\
\bottomrule
\end{tabular}
}
\caption{\textbf{Main results} on three backbone models. ``RL + Negative'' applies SFT with negative reasoning trajectories on the RL model. ``RL + P\&N'' applies DPO with chosen data from the model's correct responses and rejected data from different sources. The best results per model are in \textbf{bold}.}
\label{tab:main}
\vspace{-0.2cm}
\end{table*}

We report the main results in Table~\ref{tab:main} and summarize the key findings below.

\textbf{\ours{} consistently achieves the best average performance across all three backbone models.}
As shown in Table~\ref{tab:main}, RL + P\&N (\ours{}) obtains the highest AVG on Qwen2.5-7B-Instruct (39.89), Qwen2.5-3B-Instruct (32.69), and Llama3.1-8B-Instruct (24.77), outperforming all baselines including RL alone.
Notably, \ours{} is the \textit{only} negative sample strategy that consistently surpasses RL across all three models, confirming our hypothesis that the \textit{quality} of negative samples is the determining factor for effective preference optimization.

\textbf{\ours{} yields particularly strong gains on competition-level mathematics.}
On Qwen2.5-7B-Instruct, \ours{} improves AIME24 from 11.00 (RL) to 14.67, a relative gain of over 33\%.
On Qwen2.5-3B-Instruct, AIME25 improves from 2.67 (RL) to 4.00, and on Llama3.1-8B-Instruct, AIME25 rises from 0.33 to 2.00.
These improvements on the most challenging benchmarks suggest that training with plausible negatives sharpens the model's ability to distinguish between subtly correct and incorrect reasoning chains. This capability is critical when problems require multi-step derivations where a single subtle error can invalidate the entire solution.

\textbf{Not all negative sample strategies improve over RL.}
On Qwen2.5-7B-Instruct, all alternative negative sample methods (Negative SFT, RS, LLM Judge) yield lower AVG than RL alone (39.43). This pattern is even more pronounced on Qwen2.5-3B-Instruct, where RS drastically underperforms (28.28 vs.\ 32.14 for RL), and on Llama3.1-8B-Instruct, where both RL + Negative (23.42) and LLM Judge (22.46) fall substantially below RL (24.53).
These results indicate that naively incorporating negative samples can be counterproductive: rejection-sampled negatives are often too easily distinguishable to provide meaningful contrastive signal, while LLM-judge-based approaches lack the fine-grained reward design needed to produce consistently plausible negatives.
In contrast, our \ours{} Reward, which combines accuracy inversion, format compliance, RM-based quality assessment, and CoT coherence monitoring, generates negative samples that share similar reasoning patterns with correct solutions, forcing the model to develop finer-grained discrimination during DPO.

\textbf{\ours{} improves out-of-domain generalization.}
On Qwen2.5-7B-Instruct, \ours{} achieves an ARC score of 90.44, substantially higher than RL (85.67) and all other baselines.
On Llama3.1-8B-Instruct, \ours{} achieves the highest GPQA score (31.31) among all methods.
This demonstrates that learning to distinguish plausible-but-incorrect reasoning from correct reasoning enhances the model's general reasoning capabilities beyond the mathematical domain used during training, as the discriminative skills acquired through preference optimization transfer across tasks.

\subsection{Ablation Studies}
\label{subsec: ablation studies}

To understand the contribution of each component in the \ours{} Reward, we conduct ablation studies on Qwen2.5-7B-Instruct by removing individual reward components from the full \ours{} framework and applying the resulting negative samples via DPO directly on the base model.

\setlength\tabcolsep{5.7pt}
\begin{table*}[t]
\centering
\def\arraystretch{.99}
\setlength{\tabcolsep}{0.42em}
\resizebox{1.0\linewidth}{!}{
\begin{tabular}{ll BBBBB R dd R}
\toprule
\multirow{2}{*}{\textbf{Model}} & \multirow{2}{*}{\textbf{Setting}}
& \multicolumn{5}{c}{\textbf{In-Domain}} & \cellcolor{white}{\multirow{2}{*}{\textbf{IND AVG}}} & \multicolumn{2}{c}{\textbf{Out-of-Domain}} & \cellcolor{white}{\multirow{2}{*}{\textbf{OOD AVG}}} \\
\cmidrule(lr){3-7} \cmidrule(lr){9-10}
& & \cellcolor{white}{MATH500} & \cellcolor{white}{AIME24} & \cellcolor{white}{AIME25} & \cellcolor{white}{Olympiad} & \cellcolor{white}{AMC} & \cellcolor{white} & \cellcolor{white}{ARC} & \cellcolor{white}{GPQA} & \cellcolor{white} \\
\midrule
\multirow{7}{*}{\textbf{Qwen2.5-7B-Instruct}}
& Base                        & 75.00 & 8.67  & 4.67  & 37.48 & 40.60 & 33.28 & 83.02 & 32.83 & 57.93 \\
\cdashlinelr{2-11}
& Base + \ours{} w/o Acc      & 66.40 & 5.00  & 2.33  & 31.11 & 30.12 & 26.99 & 70.22 & 29.29 & 49.76 \\
& Base + \ours{} w/o Format   & 69.20 & 10.00 & 3.33  & 36.00 & 39.16 & 31.54 & 82.00 & 29.29 & 55.65 \\
& Base + \ours{} w/o RM       & 73.20 & 9.00  & 5.33  & 37.19 & 36.87 & 32.32 & 86.52 & 32.32 & 59.42 \\
& Base + \ours{} w/o Bucket RM & 73.60 & 8.33  & 4.67  & 36.89 & 38.19 & 32.34 & 88.57 & 30.30 & 59.44 \\
\cdashlinelr{2-11}
& Base + \ours{}              & 74.60 & 10.67 & 7.67 & 38.37 & 41.33 & \textbf{34.53} & 88.31 & 35.35 & \textbf{61.83} \\
\bottomrule
\end{tabular}
}
\caption{\textbf{Ablation studies} on Qwen2.5-7B-Instruct. We remove individual components from the \ours{} Reward and evaluate the resulting negative samples via DPO on the base model. ``w/o Acc'' removes the accuracy constraint, allowing correct answers in the negative samples. ``w/o Bucket RM'' uses raw RM scores without bucketing. The best results are in \textbf{bold}.}
\label{tab:ablation}
\vspace{-0.2cm}
\end{table*}

Table~\ref{tab:ablation} reports the results. We highlight the following observations.

\textbf{The accuracy constraint is the most critical component.}
Removing the accuracy constraint (w/o Acc) causes the largest performance drop, with IND AVG falling from 34.53 to 26.99 and OOD AVG from 61.83 to 49.76, both substantially below the Base model (33.28 and 57.93).
Without the accuracy constraint in the \ours{} Reward (\S\ref{subsubsec:rl}), the reverse RL objective cannot function as intended: the \ours{} Model is no longer steered toward producing \textit{incorrect} answers, and may instead generate correct responses that provide no contrastive signal during DPO training. This effectively collapses the preference learning, as the model cannot learn to discriminate between correct and flawed reasoning when both chosen and rejected samples may contain correct answers.

\textbf{Format compliance serves as a critical gating mechanism.}
Removing the format score (w/o Format) leads to the second-largest degradation, with IND AVG dropping to 31.54 and OOD AVG to 55.65.
As described in \S\ref{subsubsec:rl}, the Hybrid Format Score gates the entire \ours{} Reward: responses with $r_{\text{format}} = 0$ receive a penalty of $-1$ regardless of other scores. Without this gate, the \ours{} Model can exploit format violations (such as omitting the boxed answer or producing malformed tags) to trivially achieve incorrect outputs without engaging in genuine reasoning, a form of reward hacking that yields structurally degenerate negative samples unsuitable for preference learning.

\textbf{The RM score and CoT score provide complementary quality signals.}
Removing the RM score (w/o RM) reduces IND AVG from 34.53 to 32.32, while removing the CoT score yields similar degradation.
The RM score (\S\ref{subsubsec:rl}) provides a learned assessment of mathematical reasoning quality, guiding the \ours{} Model toward generating responses that are convincing to quality evaluators despite arriving at incorrect conclusions. Without it, the negative samples may contain obviously flawed reasoning that fails to challenge the target model during DPO.

\textbf{Bucketing the RM score improves training stability.}
Replacing the bucketed RM score with raw continuous scores (w/o Bucket RM) degrades IND AVG from 34.53 to 32.34, a drop comparable to removing the RM entirely. As described in \S\ref{subsubsec:rl}, the bucketing mechanism discretizes the reward signal into coarse intervals, reducing sensitivity to minor score fluctuations. Without bucketing, the continuous RM signal provides overly fine-grained gradients that the \ours{} Model can exploit during late-stage RL training, leading to reward hacking where the model learns to maximize the RM score without genuinely improving reasoning quality.

\section{Analysis}
\subsection{Judgment Capability of Reward Models}
\label{subsec: judgement capability of reward models}

To validate that our trained RM provides reliable quality signals, we construct a judgment benchmark from MATH-500 by pairing correct and incorrect responses sampled from Qwen2.5-7B-Instruct, and evaluate whether each model can identify the correct one.

\begin{figure}[t]
\centering
\includegraphics[width=0.6\linewidth]{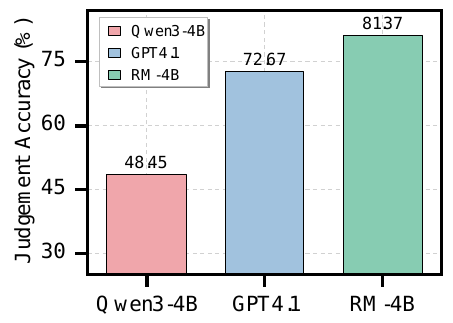}
\vspace{-0.1cm}
\caption{\textbf{Judgment accuracy on MATH-500 paired responses.} Our RM-4B substantially outperforms both Qwen3-4B and GPT-4.1.}
\label{fig:rm_capability}
\vspace{-0.2cm}
\end{figure}

\textbf{Our RM-4B substantially outperforms both Qwen3-4B and GPT-4.1 in judgment accuracy.}
As shown in Figure~\ref{fig:rm_capability}, Qwen3-4B achieves only 48.45\% accuracy (near random chance), and GPT-4.1 reaches 72.67\%. Our RM-4B achieves 81.37\%, surpassing GPT-4.1 by 8.70 points. This confirms that the CenterBT objective (Eq.~\ref{eq:bt_loss}) effectively transforms a weak base model into a strong quality evaluator for the \ours{} Reward.

\subsection{Quality of Plausible Negative Samples}
\label{subsec: quality of plausible negative samples}

We score Correct Samples (CS), Plausible Negative Samples (\ours{}), and Rejection Sampling Samples (RS) with RM-4B, and compute the Wasserstein Distance (WD) between CS and each negative sample type.

\begin{figure}[t]
\centering
\includegraphics[width=0.85\linewidth]{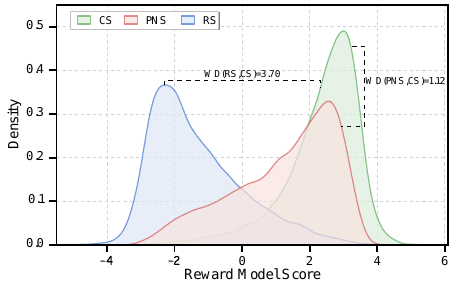}
\vspace{-0.1cm}
\caption{\textbf{Reward score distributions of different sample types.} \ours{} closely aligns with CS (WD\,=\,1.12), while RS concentrates in the negative range (WD\,=\,3.70).}
\label{fig:quality_pns}
\vspace{-0.2cm}
\end{figure}

\textbf{\ours{} achieves reasoning quality closely aligned with correct samples.}
As shown in Figure~\ref{fig:quality_pns}, RS scores concentrate in the negative range, while \ours{} scores are predominantly positive and closely aligned with CS (WD\,=\,1.12 vs.\ 3.70). This confirms that our reverse RL training produces negative samples with high reasoning quality despite incorrect conclusions. The close alignment between \ours{} and CS explains their effectiveness for DPO: they force the model to learn fine-grained distinctions rather than reject obviously flawed responses.

\subsection{Pattern Statistics of Plausible Negative Samples}
\label{subsec: pattern statistics of plausible negative samples}

We further analyze the error types in \ours{} and RS using Doubao-Seed-1.6 as an automated annotator (see Appendix~\ref{sec:error_classification_prompt}). Following the taxonomy proposed by~\citet{tian2026learning}, errors are classified into the categories in Table~\ref{tab:error_taxonomy}.

\begin{table}[t]
\centering
\small
\setlength{\tabcolsep}{0.4em}
\begin{tabular}{ll}
\toprule
\textbf{Primary Category} & \textbf{Sub-categories} \\
\midrule
Understanding   & Problem / Conceptual Misunderstanding \\
Knowledge       & Factual / Theorem / Definition Error \\
Logical         & Strategy / Reasoning / Premise / \\
                & Consistency Error \\
Calculation     & Numerical / Formula / Parameter / \\
                & Unit Error \\
Programming     & Syntax / Function / Data Type Error \\
Formal          & Symbol / Formatting Error \\
Completeness    & Boundary Omission \\
Special Cases   & Reflection / Summary Error, \\
                & Hallucination, Redundancy \\
Evaluation      & Incorrect Ground Truth, \\
                & Answer Parsing Error \\
\bottomrule
\end{tabular}
\caption{\textbf{Error taxonomy} for classifying error patterns in negative samples.}
\label{tab:error_taxonomy}
\vspace{-0.2cm}
\end{table}

\begin{figure}[h]
\centering
\includegraphics[width=0.8\linewidth]{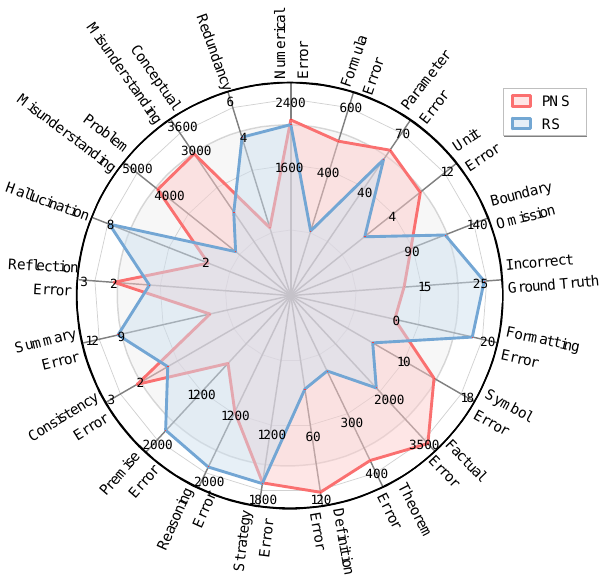}
\vspace{-0.1cm}
\caption{\textbf{Error pattern distributions of \ours{} and RS.} \ours{} errors concentrate in Knowledge Errors, while RS errors are dominated by Logical Errors.}
\label{fig:pattern_pns}
\vspace{-0.3cm}
\end{figure}

\textbf{\ours{} and RS exhibit fundamentally different error patterns.}
As shown in Figure~\ref{fig:pattern_pns}, RS errors are dominated by Logical Errors (Reasoning Error, Premise Error), which correspond to obvious logical flaws that are easy to detect during preference optimization. In contrast, \ours{} errors concentrate in Knowledge Errors (Factual Error, Theorem Error, Definition Error), where the model applies incorrect theorems or relies on wrong facts within an otherwise coherent reasoning chain. Such knowledge-level mistakes are substantially harder to identify, as they preserve the overall logical structure while introducing subtle factual deviations. This explains why \ours{} yields stronger improvements: models develop deeper mathematical understanding to distinguish responses that differ only in the correctness of specific factual claims, rather than in the coherence of their reasoning.

\section{Conclusion}

In this paper, we propose Plausible Negative Samples (\ours{}), a method that synthesizes high-quality negative samples exhibiting coherent reasoning processes yet arriving at incorrect conclusions. \ours{} trains a dedicated model via reverse reinforcement learning guided by a composite reward combining format compliance, answer accuracy inversion, reward model assessment, and chain-of-thought quality evaluation. Experiments across three backbone models and seven benchmarks show that \ours{} consistently outperforms all baselines, while naive negative sample strategies often degrade performance below the RL baseline. Our analysis further reveals that plausible negative samples closely align with correct solutions in reasoning quality and are dominated by subtle knowledge errors rather than obvious logical flaws, forcing models to develop finer-grained discrimination between sound and subtly flawed reasoning.



\section*{Impact Statement}


This paper presents work whose goal is to advance the field of Machine
Learning. There are many potential societal consequences of our work, none
which we feel must be specifically highlighted here.



\bibliography{main}

\begin{thebibliography}{46}
\providecommand{\natexlab}[1]{#1}
\providecommand{\url}[1]{\texttt{#1}}
\expandafter\ifx\csname urlstyle\endcsname\relax
  \providecommand{\doi}[1]{doi: #1}\else
  \providecommand{\doi}{doi: \begingroup \urlstyle{rm}\Url}\fi

\bibitem[Alazraki et~al.(2025)Alazraki, Mozes, Campos, Yi-Chern, Rei, and Bartolo]{alazraki-etal-2025-need}
Alazraki, L., Mozes, M., Campos, J.~A., Yi-Chern, T., Rei, M., and Bartolo, M.
\newblock No need for explanations: {LLM}s can implicitly learn from mistakes in-context.
\newblock In \emph{Proceedings of the 2025 Conference on Empirical Methods in Natural Language Processing}, pp.\  33191--33215, 2025.
\newblock ISBN 979-8-89176-332-6.
\newblock \doi{10.18653/v1/2025.emnlp-main.1686}.

\bibitem[An et~al.(2023)An, Ma, Lin, Zheng, Lou, and Chen]{an2023learning}
An, S., Ma, Z., Lin, Z., Zheng, N., Lou, J.-G., and Chen, W.
\newblock Learning from mistakes makes llm better reasoner.
\newblock \emph{arXiv preprint arXiv:2310.20689}, 2023.

\bibitem[Bradley \& Terry(1952)Bradley and Terry]{bradley1952rank}
Bradley, R.~A. and Terry, M.~E.
\newblock Rank analysis of incomplete block designs: I. the method of paired comparisons.
\newblock \emph{Biometrika}, 39\penalty0 (3/4):\penalty0 324--345, 1952.

\bibitem[Clark et~al.(2018)Clark, Cowhey, Etzioni, Khot, Sabharwal, Schoenick, and Tafjord]{clark2018think}
Clark, P., Cowhey, I., Etzioni, O., Khot, T., Sabharwal, A., Schoenick, C., and Tafjord, O.
\newblock Think you have solved question answering? try arc, the ai2 reasoning challenge.
\newblock \emph{arXiv preprint arXiv:1803.05457}, 2018.

\bibitem[Cui et~al.(2025)Cui, Yuan, Wang, Wang, Zhang, Chen, Li, He, Fan, Yu, et~al.]{cui2025prime}
Cui, G., Yuan, L., Wang, Z., Wang, H., Zhang, Y., Chen, J., Li, W., He, B., Fan, Y., Yu, T., et~al.
\newblock Process reinforcement through implicit rewards.
\newblock \emph{arXiv preprint arXiv:2502.01456}, 2025.

\bibitem[Dubey et~al.(2024)Dubey, Jauhri, Pandey, Kadian, Al-Dahle, Letman, Mathur, Schelten, Yang, Fan, et~al.]{dubey2024llama}
Dubey, A., Jauhri, A., Pandey, A., Kadian, A., Al-Dahle, A., Letman, A., Mathur, A., Schelten, A., Yang, A., Fan, A., et~al.
\newblock The llama 3 herd of models.
\newblock \emph{arXiv e-prints}, pp.\  arXiv--2407, 2024.

\bibitem[Gao \& Das(2024)Gao and Das]{gao2024contrastive}
Gao, X. and Das, K.
\newblock Customizing language model responses with contrastive in-context learning.
\newblock In \emph{Proceedings of the aaai conference on artificial intelligence}, volume~38, pp.\  18039--18046, 2024.

\bibitem[Guo et~al.(2025)Guo, Yang, Zhang, Song, Zhang, Xu, Zhu, Ma, Wang, Bi, et~al.]{guo2025deepseek}
Guo, D., Yang, D., Zhang, H., Song, J., Zhang, R., Xu, R., Zhu, Q., Ma, S., Wang, P., Bi, X., et~al.
\newblock Deepseek-r1: Incentivizing reasoning capability in llms via reinforcement learning.
\newblock \emph{arXiv preprint arXiv:2501.12948}, 2025.

\bibitem[Hamdan \& Yuret(2025)Hamdan and Yuret]{hamdan2025how}
Hamdan, S. and Yuret, D.
\newblock How much do llms learn from negative examples?
\newblock \emph{arXiv preprint arXiv:2503.14391}, 2025.

\bibitem[He et~al.(2024)He, Luo, Bai, Hu, Thai, Shen, Hu, Han, Huang, Zhang, et~al.]{he2024olympiadbench}
He, C., Luo, R., Bai, Y., Hu, S., Thai, Z.~L., Shen, J., Hu, J., Han, X., Huang, Y., Zhang, Y., et~al.
\newblock Olympiadbench: A challenging benchmark for promoting agi with olympiad-level bilingual multimodal scientific problems.
\newblock \emph{arXiv preprint arXiv:2402.14008}, 2024.

\bibitem[Hendrycks et~al.()Hendrycks, Burns, Kadavath, Arora, Basart, Tang, Song, and Steinhardt]{hendrycks2measuring}
Hendrycks, D., Burns, C., Kadavath, S., Arora, A., Basart, S., Tang, E., Song, D., and Steinhardt, J.
\newblock Measuring mathematical problem solving with the math dataset.
\newblock In \emph{Thirty-fifth Conference on Neural Information Processing Systems Datasets and Benchmarks Track (Round 2)}.

\bibitem[Holroyd \& Coles(2002)Holroyd and Coles]{LLM_COG_holroyd2002ern}
Holroyd, C.~B. and Coles, M. G.~H.
\newblock The neural basis of human error processing: reinforcement learning, dopamine, and the error-related negativity.
\newblock \emph{Psychological Review}, 109\penalty0 (4):\penalty0 679--709, 2002.

\bibitem[Hu et~al.(2025)Hu, Zhang, Han, Jiang, Zhang, and Shum]{hu2025openreasonerzero}
Hu, J., Zhang, Y., Han, Q., Jiang, D., Zhang, X., and Shum, H.-Y.
\newblock Open-reasoner-zero: An open source approach to scaling up reinforcement learning on the base model.
\newblock In \emph{The Thirty-ninth Annual Conference on Neural Information Processing Systems}, 2025.

\bibitem[Jaech et~al.(2024)Jaech, Kalai, Lerer, Richardson, El-Kishky, Low, Helyar, Madry, Beutel, Carney, et~al.]{jaech2024openai}
Jaech, A., Kalai, A., Lerer, A., Richardson, A., El-Kishky, A., Low, A., Helyar, A., Madry, A., Beutel, A., Carney, A., et~al.
\newblock Openai o1 system card.
\newblock \emph{arXiv preprint arXiv:2412.16720}, 2024.

\bibitem[Khaki et~al.(2024)Khaki, Li, Ma, Yang, and Ramachandra]{khaki-etal-2024-rs}
Khaki, S., Li, J., Ma, L., Yang, L., and Ramachandra, P.
\newblock {RS}-{DPO}: A hybrid rejection sampling and direct preference optimization method for alignment of large language models.
\newblock In \emph{Findings of the Association for Computational Linguistics: NAACL 2024}, pp.\  1665--1680. Association for Computational Linguistics, 2024.
\newblock \doi{10.18653/v1/2024.findings-naacl.108}.

\bibitem[Kumar et~al.(2010)Kumar, Packer, and Koller]{LLM_kumar2010selfpaced}
Kumar, M.~P., Packer, B., and Koller, D.
\newblock Self-paced learning for latent variable models.
\newblock In \emph{Advances in Neural Information Processing Systems (NeurIPS)}, 2010.

\bibitem[Li et~al.(2024)Li, Beeching, Tunstall, Lipkin, Soletskyi, Huang, Rasul, Yu, Jiang, Shen, et~al.]{li2024numinamath}
Li, J., Beeching, E., Tunstall, L., Lipkin, B., Soletskyi, R., Huang, S., Rasul, K., Yu, L., Jiang, A.~Q., Shen, Z., et~al.
\newblock Numinamath: The largest public dataset in ai4maths with 860k pairs of competition math problems and solutions.
\newblock \emph{Hugging Face repository}, 13\penalty0 (9):\penalty0 9, 2024.

\bibitem[Liu et~al.(2025)Liu, Liang, He, Xu, Wang, He, Tu, Mi, and Yu]{liu2025rise}
Liu, X., Liang, T., He, Z., Xu, J., Wang, W., He, P., Tu, Z., Mi, H., and Yu, D.
\newblock Trust, but verify: A self-verification approach to reinforcement learning with verifiable rewards.
\newblock In \emph{Advances in Neural Information Processing Systems (NeurIPS)}, 2025.

\bibitem[Luo et~al.(2024)Luo, Liu, Liu, Phatale, Lara, Li, Shu, Zhu, Meng, Sun, and Rastogi]{LLM_luo2024omegaprm}
Luo, L., Liu, Y., Liu, R., Phatale, S., Lara, H., Li, Y., Shu, L., Zhu, Y., Meng, L., Sun, J., and Rastogi, A.
\newblock Improve mathematical reasoning in language models by automated process supervision.
\newblock \emph{arXiv preprint arXiv:2406.06592}, 2024.

\bibitem[Ouyang et~al.(2022)Ouyang, Wu, Jiang, Almeida, Wainwright, Mishkin, Zhang, Agarwal, Slama, Ray, et~al.]{ouyang2022training}
Ouyang, L., Wu, J., Jiang, X., Almeida, D., Wainwright, C., Mishkin, P., Zhang, C., Agarwal, S., Slama, K., Ray, A., et~al.
\newblock Training language models to follow instructions with human feedback.
\newblock \emph{Advances in neural information processing systems}, 35:\penalty0 27730--27744, 2022.

\bibitem[Pan et~al.(2025)Pan, Li, Lin, Pei, Tang, Wu, Ming, Zhao, He, and Wu]{pan2025lemma}
Pan, Z., Li, Y., Lin, H., Pei, Q., Tang, Z., Wu, W., Ming, C., Zhao, H.~V., He, C., and Wu, L.
\newblock {LEMMA}: Learning from errors for {M}athe{M}atical advancement in {LLM}s.
\newblock In Che, W., Nabende, J., Shutova, E., and Pilehvar, M.~T. (eds.), \emph{Findings of the Association for Computational Linguistics: ACL 2025}, pp.\  11615--11639, Vienna, Austria, July 2025. Association for Computational Linguistics.
\newblock ISBN 979-8-89176-256-5.
\newblock \doi{10.18653/v1/2025.findings-acl.605}.

\bibitem[Quamar et~al.(2025)Quamar, Areeb, Kuznetsov, Ozmen, and Celik]{LLM_quamar2025stars}
Quamar, M.~A., Areeb, M., Kuznetsov, M., Ozmen, M.~O., and Celik, Z.~B.
\newblock Stars: Segment-level token alignment with rejection sampling in large language models.
\newblock \emph{arXiv preprint arXiv:2511.03827}, 2025.

\bibitem[Qwen et~al.(2025)Qwen, Yang, Yang, Zhang, et~al.]{qwen2025qwen25technicalreport}
Qwen, Yang, A., Yang, B., Zhang, B., et~al.
\newblock Qwen2.5 technical report, 2025.

\bibitem[Rafailov et~al.(2023)Rafailov, Sharma, Mitchell, Manning, Ermon, and Finn]{rafailov2023direct}
Rafailov, R., Sharma, A., Mitchell, E., Manning, C.~D., Ermon, S., and Finn, C.
\newblock Direct preference optimization: Your language model is secretly a reward model.
\newblock \emph{Advances in neural information processing systems}, 36:\penalty0 53728--53741, 2023.

\bibitem[Rein et~al.()Rein, Hou, Stickland, Petty, Pang, Dirani, Michael, and Bowman]{rein2024gpqa}
Rein, D., Hou, B.~L., Stickland, A.~C., Petty, J., Pang, R.~Y., Dirani, J., Michael, J., and Bowman, S.~R.
\newblock Gpqa: A graduate-level google-proof q\&a benchmark.
\newblock In \emph{First Conference on Language Modeling}.

\bibitem[Shao et~al.(2024{\natexlab{a}})Shao, Wang, Zhu, Xu, Song, Bi, Zhang, Zhang, Li, Wu, and Guo]{LLM_shao2024deepseekmath}
Shao, Z., Wang, P., Zhu, Q., Xu, R., Song, J., Bi, X., Zhang, H., Zhang, M., Li, Y.~K., Wu, Y., and Guo, D.
\newblock Deepseekmath: Pushing the limits of mathematical reasoning in open language models.
\newblock \emph{arXiv preprint arXiv:2402.03300}, 2024{\natexlab{a}}.

\bibitem[Shao et~al.(2024{\natexlab{b}})Shao, Wang, Zhu, Xu, Song, Bi, Zhang, Zhang, Li, Wu, and Guo]{shao2024deepseekmath}
Shao, Z., Wang, P., Zhu, Q., Xu, R., Song, J., Bi, X., Zhang, H., Zhang, M., Li, Y.~K., Wu, Y., and Guo, D.
\newblock Deepseekmath: Pushing the limits of mathematical reasoning in open language models, 2024{\natexlab{b}}.

\bibitem[Sheng et~al.(2025)Sheng, Zhang, Ye, Wu, Zhang, Zhang, Peng, Lin, and Wu]{sheng2025hybridflow}
Sheng, G., Zhang, C., Ye, Z., Wu, X., Zhang, W., Zhang, R., Peng, Y., Lin, H., and Wu, C.
\newblock Hybridflow: A flexible and efficient rlhf framework.
\newblock In \emph{Proceedings of the Twentieth European Conference on Computer Systems}, pp.\  1279--1297, 2025.

\bibitem[Team et~al.(2025)Team, Du, Gao, Xing, Jiang, Chen, Li, Xiao, Du, Liao, et~al.]{team2025kimi}
Team, K., Du, A., Gao, B., Xing, B., Jiang, C., Chen, C., Li, C., Xiao, C., Du, C., Liao, C., et~al.
\newblock Kimi k1. 5: Scaling reinforcement learning with llms.
\newblock \emph{arXiv preprint arXiv:2501.12599}, 2025.

\bibitem[Tian et~al.(2026)Tian, Ma, Xu, Lyu, Li, Dong, Chu, Wang, and Shen]{tian2026learning}
Tian, X., Ma, M., Xu, B., Lyu, N., Li, W., Dong, H., Chu, Z., Wang, Y., and Shen, H.
\newblock Learning from mistakes: Negative reasoning samples enhance out-of-domain generalization.
\newblock \emph{arXiv preprint arXiv:2601.04992}, 2026.

\bibitem[Tong et~al.(2024)Tong, Li, Wang, Wang, Teng, and Shang]{tong2024can}
Tong, Y., Li, D., Wang, S., Wang, Y., Teng, F., and Shang, J.
\newblock Can {LLM}s learn from previous mistakes? investigating {LLM}s' errors to boost for reasoning.
\newblock In Ku, L.-W., Martins, A., and Srikumar, V. (eds.), \emph{Proceedings of the 62nd Annual Meeting of the Association for Computational Linguistics (Volume 1: Long Papers)}, pp.\  3065--3080, Bangkok, Thailand, August 2024. Association for Computational Linguistics.
\newblock \doi{10.18653/v1/2024.acl-long.169}.

\bibitem[Touvron et~al.(2023)Touvron, Martin, Stone, Albert, Almahairi, Babaei, Bashlykov, Batra, Bhargava, Bhosale, et~al.]{touvron2023llama2}
Touvron, H., Martin, L., Stone, K., Albert, P., Almahairi, A., Babaei, Y., Bashlykov, N., Batra, S., Bhargava, P., Bhosale, S., et~al.
\newblock Llama 2: Open foundation and fine-tuned chat models.
\newblock \emph{arXiv preprint arXiv:2307.09288}, 2023.

\bibitem[Wang et~al.(2024)Wang, Li, Han, Zhang, and Baldwin]{wang2024learning}
Wang, R., Li, H., Han, X., Zhang, Y., and Baldwin, T.
\newblock Learning from failure: Integrating negative examples when fine-tuning large language models as agents.
\newblock \emph{arXiv preprint arXiv:2402.11651}, 2024.

\bibitem[Wei et~al.(2022)Wei, Bosma, Zhao, Guu, Yu, Lester, Du, Dai, and Le]{LLM_wei2022finetuned}
Wei, J., Bosma, M., Zhao, V., Guu, K., Yu, A., Lester, B., Du, N., Dai, A., and Le, Q.
\newblock Finetuned language models are zero-shot learners.
\newblock \emph{arXiv preprint arXiv:2109.01652}, 2022.

\bibitem[Wen et~al.(2025)Wen, Liu, Zheng, Xu, Ye, Wu, Liang, Wang, Li, Miao, Bian, and Yang]{wen2025rlvr_reasoning}
Wen, X., Liu, Z., Zheng, S., Xu, Z., Ye, S., Wu, Z., Liang, X., Wang, Y., Li, J., Miao, Z., Bian, J., and Yang, M.
\newblock Reinforcement learning with verifiable rewards implicitly incentivizes correct reasoning in base llms.
\newblock \emph{arXiv preprint arXiv:2506.14245}, 2025.

\bibitem[Xiong et~al.(2025)Xiong, Yao, Xu, Pang, Wang, Sahoo, Li, Jiang, Zhang, Xiong, and Dong]{LLM_xiong2025minimalist}
Xiong, W., Yao, J., Xu, Y., Pang, B., Wang, L., Sahoo, D., Li, J., Jiang, N., Zhang, T., Xiong, C., and Dong, H.
\newblock A minimalist approach to {LLM} reasoning: from rejection sampling to reinforce.
\newblock \emph{arXiv preprint arXiv:2504.11343}, 2025.

\bibitem[Yang et~al.(2025)Yang, Li, Yang, Zhang, Hui, Zheng, Yu, Gao, Huang, Lv, et~al.]{yang2025qwen3}
Yang, A., Li, A., Yang, B., Zhang, B., Hui, B., Zheng, B., Yu, B., Gao, C., Huang, C., Lv, C., et~al.
\newblock Qwen3 technical report.
\newblock \emph{arXiv preprint arXiv:2505.09388}, 2025.

\bibitem[Yu et~al.(2025{\natexlab{a}})Yu, Li, Liao, Zhu, Xue, Xu, Wang, Hong, Mi, and Shang]{yu2025selferror}
Yu, E., Li, J., Liao, M., Zhu, Q., Xue, B., Xu, M., Wang, B., Hong, L., Mi, F., and Shang, L.
\newblock Self-error-instruct: Generalizing from errors for {LLM}s mathematical reasoning.
\newblock In Che, W., Nabende, J., Shutova, E., and Pilehvar, M.~T. (eds.), \emph{Proceedings of the 63rd Annual Meeting of the Association for Computational Linguistics (Volume 1: Long Papers)}, pp.\  8504--8519, Vienna, Austria, July 2025{\natexlab{a}}. Association for Computational Linguistics.
\newblock ISBN 979-8-89176-251-0.
\newblock \doi{10.18653/v1/2025.acl-long.417}.

\bibitem[Yu et~al.(2025{\natexlab{b}})Yu, Zhang, Zhu, Yuan, Zuo, Yue, Dai, Fan, Liu, Liu, et~al.]{yu2025dapo}
Yu, Q., Zhang, Z., Zhu, R., Yuan, Y., Zuo, X., Yue, Y., Dai, W., Fan, T., Liu, G., Liu, L., et~al.
\newblock Dapo: An open-source llm reinforcement learning system at scale.
\newblock \emph{arXiv preprint arXiv:2503.14476}, 2025{\natexlab{b}}.

\bibitem[Yuan et~al.(2025{\natexlab{a}})Yuan, Cui, Wang, Gao, Zhou, and Naseem]{yuan2025kardia}
Yuan, J., Cui, Z., Wang, H., Gao, Y., Zhou, Y., and Naseem, U.
\newblock Kardia-r1: Unleashing llms to reason toward understanding and empathy for emotional support via rubric-as-judge reinforcement learning.
\newblock \emph{arXiv preprint arXiv:2512.01282}, 2025{\natexlab{a}}.

\bibitem[Yuan et~al.(2025{\natexlab{b}})Yuan, Li, Chen, Cui, Ding, Zhang, Zhou, Liu, and Peng]{yuan2025free}
Yuan, L., Li, W., Chen, H., Cui, G., Ding, N., Zhang, K., Zhou, B., Liu, Z., and Peng, H.
\newblock Free process rewards without process labels.
\newblock In \emph{Forty-second International Conference on Machine Learning}, 2025{\natexlab{b}}.

\bibitem[Yue et~al.(2025)Yue, Chen, Lu, Zhao, Wang, Yue, Song, and Huang]{yue2025does}
Yue, Y., Chen, Z., Lu, R., Zhao, A., Wang, Z., Yue, Y., Song, S., and Huang, G.
\newblock Does reinforcement learning really incentivize reasoning capacity in {LLM}s beyond the base model?
\newblock In \emph{The Thirty-ninth Annual Conference on Neural Information Processing Systems}, 2025.

\bibitem[Zeng et~al.(2025)Zeng, Huang, Liu, Liu, He, Ma, and He]{zeng2025simplerl}
Zeng, W., Huang, Y., Liu, Q., Liu, W., He, K., Ma, Z., and He, J.
\newblock Simplerl-zoo: Investigating and taming zero reinforcement learning for open base models in the wild.
\newblock \emph{arXiv preprint arXiv:2503.18892}, 2025.

\bibitem[Zhang et~al.(2025)Zhang, Zheng, Wu, Zhang, Lin, Yu, Liu, Zhou, and Lin]{zhang-etal-2025-lessons}
Zhang, Z., Zheng, C., Wu, Y., Zhang, B., Lin, R., Yu, B., Liu, D., Zhou, J., and Lin, J.
\newblock The lessons of developing process reward models in mathematical reasoning.
\newblock In \emph{Findings of the Association for Computational Linguistics: ACL 2025}, pp.\  10495--10516, 2025.
\newblock ISBN 979-8-89176-256-5.
\newblock \doi{10.18653/v1/2025.findings-acl.547}.

\bibitem[Zhao et~al.(2024)Zhao, Huang, Hu, Wang, Mao, Zhang, Jiang, Wu, Ai, Wang, Zhou, and Chen]{zhao2024swift}
Zhao, Y., Huang, J., Hu, J., Wang, X., Mao, Y., Zhang, D., Jiang, Z., Wu, Z., Ai, B., Wang, A., Zhou, W., and Chen, Y.
\newblock Swift: A scalable lightweight infrastructure for fine-tuning, 2024.

\bibitem[Zhu et~al.(2025)Zhu, Wang, Sun, Chen, Liu, Zhang, and Wang]{zhu2025continual}
Zhu, K., Wang, Y., Sun, Y., Chen, Q., Liu, J., Zhang, G., and Wang, J.
\newblock Continual sft matches multimodal rlhf with negative supervision.
\newblock In \emph{Proceedings of the Computer Vision and Pattern Recognition Conference}, pp.\  14615--14624, 2025.

\end{thebibliography}
\bibliographystyle{icml2026}

\newpage
\appendix
\onecolumn
\section{Experiments Details}
\label{sec:exp_details}
\paragraph{\textbf{Hardware and Software Platform.}}
RM training and DPO training are conducted on servers equipped with 16 NVIDIA A800-SXM4 GPUs, using the MS-SWIFT framework~\citep{zhao2024swift}. RL training is performed on servers with 32 NVIDIA H20 GPUs, using the VeRL framework~\citep{sheng2025hybridflow}.

\paragraph{\textbf{RM Training Configuration.}}
The reward model is trained with full-parameter optimization for one epoch. The maximum sequence length is set to 24,000 tokens. Training uses a per-device batch size of 1 with gradient accumulation over 8 steps. The model is optimized with Adam using a learning rate of $1 \times 10^{-5}$, together with a cosine learning rate schedule and a warmup ratio of 0.1. We enable bf16 precision and employ DeepSpeed ZeRO-2 for efficient training.

\paragraph{\textbf{RL Training Configuration.}}
We train the \ours{} Model using the GRPO algorithm for 200 optimization steps. During training, rollouts are generated using temperature sampling with $\tau = 1.0$, and end-of-sequence tokens are enforced. We adopt a grouped sampling strategy with a group size of $G = 8$ responses per problem. The maximum prompt length is set to 24,000 tokens, and the maximum response length is 8,000 tokens. The actor model is optimized using Adam with a learning rate of $1 \times 10^{-6}$ and a batch size of 128. The clipping range is set to $[0.01, 0.99]$. We set the reward weighting coefficients $\lambda_r = \lambda_c = 0.5$ by default.

\paragraph{\textbf{DPO Training Configuration.}}
DPO training is performed with full-parameter optimization for one epoch. The maximum sequence length is set to 24,000 tokens. Training uses a per-device batch size of 1. The model is optimized with Adam using a learning rate of $1 \times 10^{-5}$, together with a cosine learning rate schedule and a warmup ratio of 0.1. We employ DeepSpeed ZeRO-2 for parallel training.

\paragraph{\textbf{Inference Configuration.}}
During inference, we use vLLM for efficient decoding. The sampling temperature is set to 0 with top-p sampling at 0.95. The maximum generation length is 24,000 tokens. For AIME24, AIME25, and AMC evaluations, the sampling temperature is set to 0.6.

\section{Prompt Templates}

We provide the prompt templates used in our experiments, including:
\begin{itemize}[leftmargin=1.5em, topsep=2pt, itemsep=1pt]
\item The system prompt used for mathematical reasoning tasks.
\item The prompt for LLM-as-a-Judge ($r_{\text{judge}}$) in the Hybrid Format Score, which evaluates whether a response contains adequate step-by-step mathematical derivations.
\item The prompt for the CoT Score ($r_{\text{cot}}$) that assesses reasoning quality across four dimensions.
\item The prompt for error classification used in our error pattern analysis (\S\ref{subsec: pattern statistics of plausible negative samples}).
\end{itemize}

\begin{tcolorbox}[title=System Prompt for Mathematical Reasoning, colback=white, colframe=black!75!white, breakable]
\begin{spacing}{1.1}
\label{sec:system_prompt}
You are a helpful AI Assistant that provides well-reasoned and detailed responses. You should FIRST output your internal thought process and then provide the final answer. The reasoning process MUST BE enclosed within \texttt{<think>} \texttt{</think>} tags. The final answer MUST BE put in \texttt{\textbackslash boxed\{\}}.
\end{spacing}
\end{tcolorbox}

\begin{tcolorbox}[title=Prompt for Hybrid Format Score ($r_{\text{judge}}$), colback=white, colframe=black!75!white, breakable]
\begin{spacing}{1.1}
\label{sec:judge_prompt}
You are a strict, stable, and conservative judge for math solution quality.

\textbf{Your job:} Decide whether the given ``model response'' contains an ADEQUATE step-by-step mathematical solving process.

\textbf{Definition of ADEQUATE step-by-step solving process (PASS):}
\begin{itemize}[leftmargin=1.5em, topsep=2pt, itemsep=1pt]
\item Contains explicit, checkable mathematical derivations or computations, such as: equation setup and transformations; substitutions, expansions, simplifications; intermediate numeric calculations; long multiplication / division / remainder steps; clearly shown intermediate results that lead to the final answer.
\item The reasoning must be more than a high-level plan; it must show actual math work.
\end{itemize}

\textbf{FAIL if any of these dominate:}
\begin{itemize}[leftmargin=1.5em, topsep=2pt, itemsep=1pt]
\item Only ``planning'' language (e.g., ``we will use theorem X'') without actually applying it with concrete steps.
\item Jumps to the final answer with little/no intermediate math.
\item Empty / meaningless / irrelevant \texttt{<think>} or text.
\item Mostly vague explanation with no checkable math.
\end{itemize}

\textbf{Output rules (VERY IMPORTANT):}
\begin{enumerate}[leftmargin=1.5em, topsep=2pt, itemsep=1pt]
\item First write a brief analysis (1--5 sentences max).
\item Then output ONLY one JSON object inside \texttt{<final>} \ldots\ \texttt{</final>}.
\item The JSON must have EXACTLY ONE key: ``verdict''
\item ``verdict'' must be exactly one of: ``pass'' or ``fail''
\item Do not output any other keys, text, or formatting outside \texttt{<final>}.
\end{enumerate}

\textbf{[model response]}\\
\texttt{\{response\}}

Output strictly like:\\
\texttt{<final>}\\
\texttt{\{"verdict":"pass"\}}\\
\texttt{</final>}
\end{spacing}
\end{tcolorbox}

\begin{tcolorbox}[title=Prompt for CoT Score ($r_{\text{cot}}$), colback=white, colframe=black!75!white, breakable]
\begin{spacing}{1.1}
\label{sec:cot_prompt}
You are a strict, stable, and conservative judge for math problems. Given the ``task problem'', score the ``model response'' on the FOUR dimensions below.\\
Each dimension must be an integer from 0 to 3:
\begin{itemize}[leftmargin=1.5em, topsep=2pt, itemsep=1pt]
\item 3: Excellent / no or almost no issues
\item 2: Minor issues but acceptable overall
\item 1: Major issues / clearly problematic
\item 0: Essentially unusable
\end{itemize}

\textbf{Four dimensions (MUST output a numeric score for each):}
\begin{enumerate}[leftmargin=1.5em, topsep=2pt, itemsep=1pt]
\item \textbf{Reasoning validity:} Does \texttt{<think>} contain steps that are directly relevant to the problem? Heavily penalize: empty \texttt{<think>}, irrelevant/nonsensical text, ``theorem/technique name-dropping'' without actually using it.
\item \textbf{Reasoning--conclusion consistency:} Does the reasoning support the conclusion? The reasoning should include checkable mathematical derivations, not merely a high-level plan. Heavily penalize: conclusion not supported by shown steps; reasoning contains only planning/intent without explicit derivation.
\item \textbf{Instruction following:} Does the response follow all constraints? Does it include off-task or disallowed extras?
\item \textbf{Repetition issues and unnecessary language mixing:} Mechanical repetition, templated filler, excessive restating. Unnecessary mixing of languages or styles that harms readability.
\end{enumerate}

\textbf{[task problem]}\\
\texttt{\{prompt\}}

\textbf{[model response]}\\
\texttt{\{response\}}

Output strictly in the following format (ONLY one JSON inside \texttt{<final>}):\\
\texttt{<final>}\\
\texttt{\{}\\
\quad\texttt{"Reasoning validity": 0-3,}\\
\quad\texttt{"Reasoning-conclusion consistency": 0-3,}\\
\quad\texttt{"Instruction following": 0-3,}\\
\quad\texttt{"Repetition issues and unnecessary language mixing": 0-3}\\
\texttt{\}}\\
\texttt{</final>}
\end{spacing}
\end{tcolorbox}

\begin{tcolorbox}[title=Prompt for Error Classification, colback=white, colframe=black!75!white, breakable]
\begin{spacing}{1.1}
\label{sec:error_classification_prompt}
You are an expert AI assistant tasked with identifying the single, most specific error category from the list below.

\textbf{Error Category List:}
\begin{itemize}[leftmargin=1.5em, topsep=2pt, itemsep=1pt]
\item \textbf{Understanding Errors:} Problem Misunderstanding, Conceptual Misunderstanding
\item \textbf{Knowledge Errors:} Factual Error, Theorem Error, Definition Error
\item \textbf{Logical Errors:} Strategy Error, Reasoning Error, Premise Error, Consistency Error
\item \textbf{Calculation Errors:} Numerical Error, Formula Error, Parameter Error, Unit Error
\item \textbf{Programming Errors:} Syntax Error, Function Error, Data Type Error
\item \textbf{Formal Errors:} Symbol Error, Formatting Error
\item \textbf{Completeness Errors:} Boundary Omission
\item \textbf{Special Cases:} Reflection Error, Summary Error, Hallucination, Redundancy
\item \textbf{Evaluation System Errors:} Incorrect Ground Truth, Correct Answer Parsing Error
\end{itemize}

\textbf{Data for Analysis:}
\begin{itemize}[leftmargin=1.5em, topsep=2pt, itemsep=1pt]
\item Question: \texttt{\{question\}}
\item Ground Truth Answer: \texttt{\{groundtruth\}}
\item Model's Reasoning Process (to be analyzed): \texttt{\{model\_reasoning\}}
\end{itemize}

\textbf{CRITICAL INSTRUCTION:}
Analyze the provided reasoning process. Your response MUST be ONLY a single, raw JSON object with the keys ``sub\_category'' and ``analysis''. Do not include any other text, explanations, apologies, or markdown formatting.

Example of a perfect response:\\
\texttt{\{"sub\_category": "Premise Error", "analysis": "The model incorrectly assumed that all bicycles use plastic squares for identification, which is a flawed premise not supported by the question's context."\}}
\end{spacing}
\end{tcolorbox}

\end{document}